\documentclass[10pt,twocolumn,letterpaper]{article}

\usepackage{wacv}
\usepackage{times}
\usepackage{epsfig}
\usepackage{graphicx}
\usepackage{amssymb}
\usepackage{times}
\usepackage{sidecap}
\usepackage{lineno,hyperref}
\usepackage[T1]{fontenc}
\usepackage{setspace}	
\usepackage{subfigure}
\usepackage{float}
\usepackage{stfloats}
\usepackage[linesnumbered,ruled,vlined]{algorithm2e}
\usepackage{adjustbox}
\usepackage{multirow}
\usepackage{hhline}
\usepackage{gensymb}
\usepackage{algorithmic}
\usepackage{amsmath,amsfonts,amsthm,bm}
\usepackage{pifont}
\usepackage{fixmath}
\usepackage{epstopdf}
\usepackage{cleveref}
\usepackage{makecell}
\usepackage{footnote}
\usepackage{color,soul}

\wacvfinalcopy 

\def\wacvPaperID{347} 

\ifwacvfinal\fi
\begin{document}

\title{Cross-Domain Face Synthesis using a Controllable GAN}

\author{Fania Mokhayeri \hspace{2cm} Kaveh Kamali \hspace{2cm} Eric Granger \\
Laboratoire d'imagerie, de vision et d'intelligence artificielle (LIVIA) \\
\'Ecole de technologie sup\'erieure, Montreal, Canada\\
}

\maketitle
\ifwacvfinal\thispagestyle{empty}\fi

\begin{abstract}
The performance of face recognition (FR) systems applied in video surveillance has been shown to improve when the design data is augmented through synthetic face generation. This is true, for instance, with pair-wise matchers (e.g., deep Siamese networks) that typically rely on a reference gallery with one still image per individual. However, generating synthetic images in the source domain may not improve the performance during operations due to the domain shift w.r.t. the target domain. Moreover, despite the emergence of Generative Adversarial Networks (GANs) for realistic synthetic generation, it is often difficult to control the conditions under which synthetic faces are generated.
In this paper, a cross-domain face synthesis approach is proposed that integrates a new Controllable GAN (C-GAN). It employs an off-the-shelf 3D face model as a simulator to generate face images under various poses. The simulated images and noise are input to the C-GAN for realism refinement which employs an additional adversarial game as a third player to preserve the identity and specific facial attributes of the refined images. This allows generating realistic synthetic face images that reflects capture conditions in the target domain while controlling the GAN output to generate faces under desired pose conditions. 
Experiments were performed using videos from the Chokepoint and COX-S2V datasets, and a deep Siamese network for FR with a single reference still per person. Results indicate that the proposed approach can provide a higher level of accuracy compared to the current state-of-the-art approaches for synthetic data augmentation\footnote{Code available at: \textcolor{red}{\href{https://github.com/faniamokhayeri/C-GAN}{\url{ https://github.com/faniamokhayeri/C-GAN}.}}}.
\end{abstract}

\section{Introduction}
\label{intro}
Recent advances in deep learning have significantly increased the performance of still-to-video face recognition (FR) systems applied in video monitoring and surveillance. One of the pioneering techniques in this area is FaceNet \cite{facenet}. It uses a deep Siamese network architecture, where the same CNN feature extractor is trained through similarity learning to perform pair-wise matching of query (video) and reference (still) faces. Despite many recent advances, FR with a single sample per person (SSPP) remains a challenging problem in video-based security and surveillance applications. In such cases, the performance of deep learning models for FR can decline significantly due to the limited robustness of matching to a single still obtained during enrolment \cite{S2}. One effective solution to alleviate the aforementioned problem is extending the gallery using synthetic face images.

Some of the recent research \cite{fania2, FML, zhao} augment galleries using synthetic images generated from 3D models. Tran et. al \cite{Tran1} proposed a face synthesis technique where CNN is employed to regress the 3D model parameters to overcome the shortage of training data. Although their results are encouraging, the synthetic face images may not be realistic enough to represent intra-class variations of target domain capture conditions. The synthetic images generated in this way are highly correlated with the original facial stills from enrolment, and there is typically a domain shift between the distribution of synthetic faces and that of faces captured in the target domain. The models naively trained on these synthetic images, often fail to generalize well when matched to real images captures in the target domain. Authors in \cite{fania} proposed an algorithm for domain-specific face synthesis (DSFS) that exploits the intra-class variation information available from the target domain.

\begin{figure*}[t]
\advance\leftskip 0.3cm
\includegraphics[width=17cm]{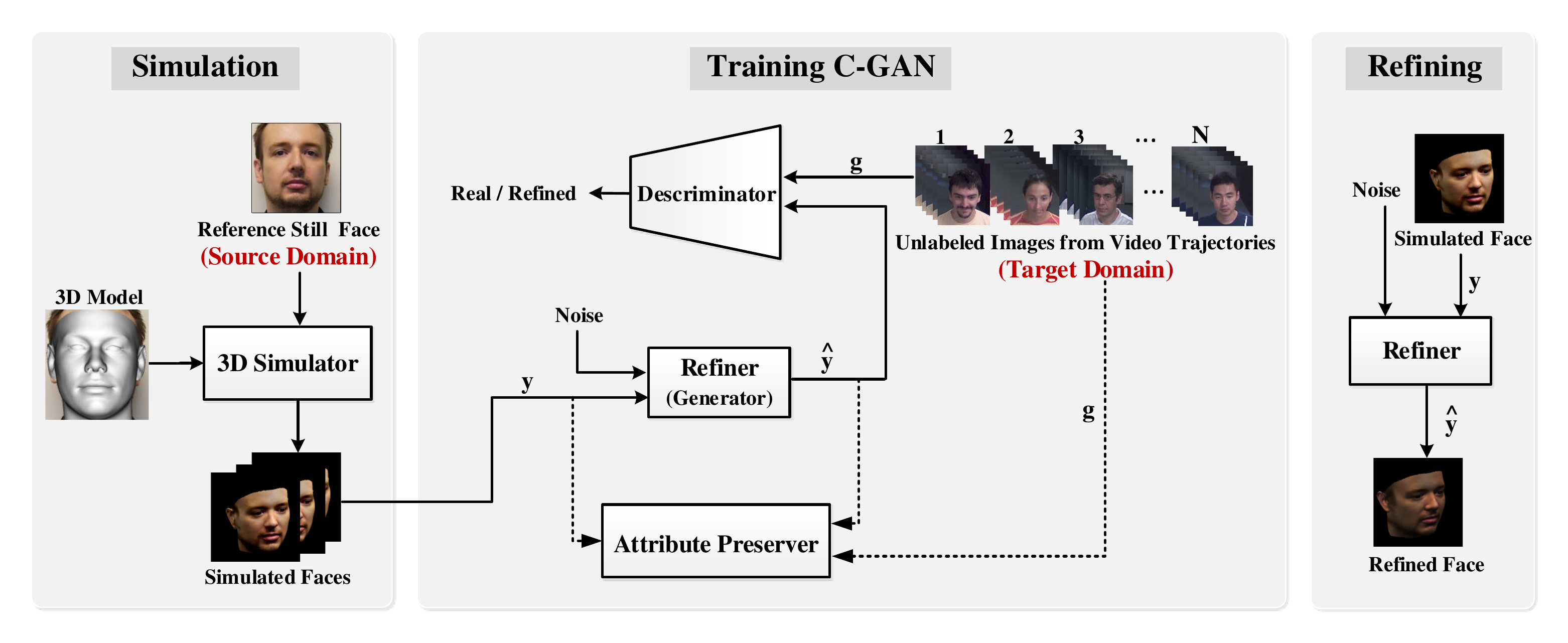}
\caption{\small An overview of the proposed cross-domain face synthesis approach based on the C-GAN. The 3D simulator generates simulated faces, $\mathbf{y}$, with the arbitrary pose. The refiner is trained using $\mathbf{y}$ the generic set, $\mathbf{g}$, and random noise to generate refined images, $\mathbf{\hat{y}}$, under the target domain capture conditions, and while specifying the pose of $\mathbf{y}$ using an additional adversarial game.}
\label{fig:1}
\end{figure*}

Generative adversarial networks (GANs) have recently shown promising results for the synthesis of  realistic face images \cite{bao2,brock,goodfellow}. For instance, DA-GAN \cite{zhao} has been proposed for automatically generating augmented data for FR in unconstrained conditions. One of the challenging issues in GAN-based face synthesis models is the difficulty of controlling images they generate since a random distribution is used as the input of generators. Modified GAN architectures, like the conditional GAN, have attempted to address this issue by setting conditions on the generative and discriminative networks for conditional image synthesis \cite{isola,lin,drgan}. However, the mapping of conditional GANs does not constrain the output to the target manifold, thus the output can be arbitrarily off the target manifold. Generating identity-preserving faces is another unsolved challenge in GAN-based face synthesis models.

In this paper, we propose a cross-domain face synthesis approach that relies on a new controllable GAN (C-GAN). It extends the original GAN by using an additional adversarial game as the third player to the GAN, competing with the refiner (generator) to preserve the specific attributes, and accordingly, providing control over the face generation process. As depicted in Figure \ref{fig:1}, C-GAN involves three main steps: (1) generating simulated face images via 3D morphable model \cite{blanz1} rendered under a specified pose, (2) refining the realism of the simulated face images using an unlabeled generic set to adapt synthetic face images in the source domain to appear as if drawn from the target domain, and (3) preserving the specific attributes of the simulated face images during the refinement through another adversarial network.
Using C-GAN, a set of realistic synthetic facial images are generated that represent gallery stills under the target domain with high consistency, while preserving their identity and allowing to specify the pose conditions of synthetic images. The refined synthetic face images are then used to augment the reference gallery to boost the performance of FR with SSPP.
The main contribution of this paper is a novel cross-domain face synthesis approach that integrates C-GAN to leverage an additional adversarial game as third player into the GAN model, producing highly consistent realistic face images in a controllable manner. Additionally, we show that using the images generated by C-GAN as additional design data within a Siamese network allows to improve still-to-video FR performance under unconstrained capture conditions.

For proof-of-concept experiments, the performance of the proposed and baseline face synthesis methods are evaluated using a "recognition via generation" framework ~\cite{zhao} on videos from the publicly available Chokepoint and COX-S2V datasets. In a particular implementation, we extend the reference gallery of a deep Siamese network for still-to-video FR.

\section{Related Work}
\label{RW}
\paragraph{GANs for Realistic Face Synthesis.}
Recently, Generative Adversarial Networks (GANs) \cite{goodfellow} have shown promising performance in face synthesis. 
Existing methods typically formulate GAN as a two-player game, where a discriminator $D$ distinguishes face images from the real and synthesized domains, while a generator $G$ reduces its discriminativeness by synthesizing a face of realistic quality. Their competition converges when the discriminator is unable to differentiate these two domains.
Benefiting from GAN, FaceID-GAN \cite{shen} is proposed which treats a classifier of face identity as the third player, competing with the generator by distinguishing the identities of the real and synthesized faces. 
The major shortcoming of the GAN-based face synthesis models is that they may produce images that are inconsistent due to the weak global constraints. To reduce this gap, Shrivastava \textit{et al.} developed SimGAN that learns a model using synthetic images as inputs instead of random noise vector~\cite{shrivastava}.  Our work draws inspiration from SimGAN \cite{shrivastava} specialized for face synthesis. 
%
Another issue of vanilla GAN is that it is difficult to control the output of the generator. Recently, conditional GANs have added condition information to the generative network and the discriminative network for conditional image synthesis \cite{isola,lin}. Tran \textit{et al.} proposed DR-GAN takes a pose code in addition to random noise vector as the inputs for discriminator with the goal of generating a face of the same identity with the target pose that can fool the discriminator \cite{drgan}. In the CAPG-GAN \cite{hu}, a couple-agent discriminator is introduced which forms a mask image to guide the generator in the learning process and provides a flexible controllable condition during inference.
The bottleneck of conditional GANs is the regression of the generator may lead to arbitrarily large errors in the output, which makes it unreliable for real-world applications \cite{chrysos}. This paper aims to address the above problems by augmenting the refiner of GAN with a domain-invariant feature extractor.

\paragraph{Domain-Invariant Representations.}
Recently, there have been efforts \cite{zhuu} to produce domain-invariant feature representations from a single input. One of the most popular approaches in this area is the domain-adversarial neural network which integrates a gradient reversal layer into the standard architecture to ensure a domain invariant feature representation \cite{ganin}. They introduced a domain confusion loss term to learn domain-invariant feature. Haeusser \textit{et al.} \cite{haeusser} produce statistically domain invariant embedding by reinforcing associations between source and target data directly in embedding space. A slightly different approach is presented in \cite{ghifary}, where common feature assimilation is achieved implicitly by using a decoder to reconstruct the input source and target images. In a similar spirit, \cite{rama} uses a generator from the encoded features to generate samples which follow the same distribution as the source dataset. 

\section{Proposed Approach} 
\label{proposed}
In the following, the set $\mathbf{X} = \{ \mathbf{x}_1, \dotsc,\mathbf{x}_i, \dotsc ,\mathbf{x}_N  \} \in \mathbb{R}^{d \times N}$ denote a gallery set composed of $n$ reference still ROIs belonging to one of $k$ different classes in the source domain, where $d$ is the number of pixels representing a ROI and $N=kn$ is the total number of reference still ROIs; 
$ \mathbf{Y} = \{ \mathbf{y}_1, \dotsc,\mathbf{y}_i, \dotsc ,\mathbf{y}_M  \} \in \mathbb{R}^{d \times M}$ and $ \mathbf{H} = \{ \mathbf{h}^1, \dotsc,\mathbf{h}^i, \dotsc ,\mathbf{h}^M  \} \in \mathbb{R}^{(kp) \times M}$ denote the simulated set and the corresponding one-hot 
labels, where $M$ is the number of the simulated ROIs and $kp$ is number of 3D simulated classes ($k$ identity class with $p$ pose). 
The label associated with $\mathbf{y}_i$ is defined as $\mathbf{h}^i=\{h^i_d,h^i_p\}$, where $h_d$ represents the label for identity and $h_p$ for pose. 
$\mathbf{G} = \{ \mathbf{g}_1, \dotsc,\mathbf{g}_i, \dotsc ,\mathbf{g}_L \} \in \mathbb{R}^{d \times L}$ denote a generic set composed of $L$ unlabeled video ROIs in the target domain.
The objective of C-GAN model is to generate realistic face images with high consistency while specifying the main attributes, in particular pose $h_p$ and identity, shown in synthetic images and preserving the identity $h_d$.

\begin{figure*}[t]
\centering 
\includegraphics[width=15cm]{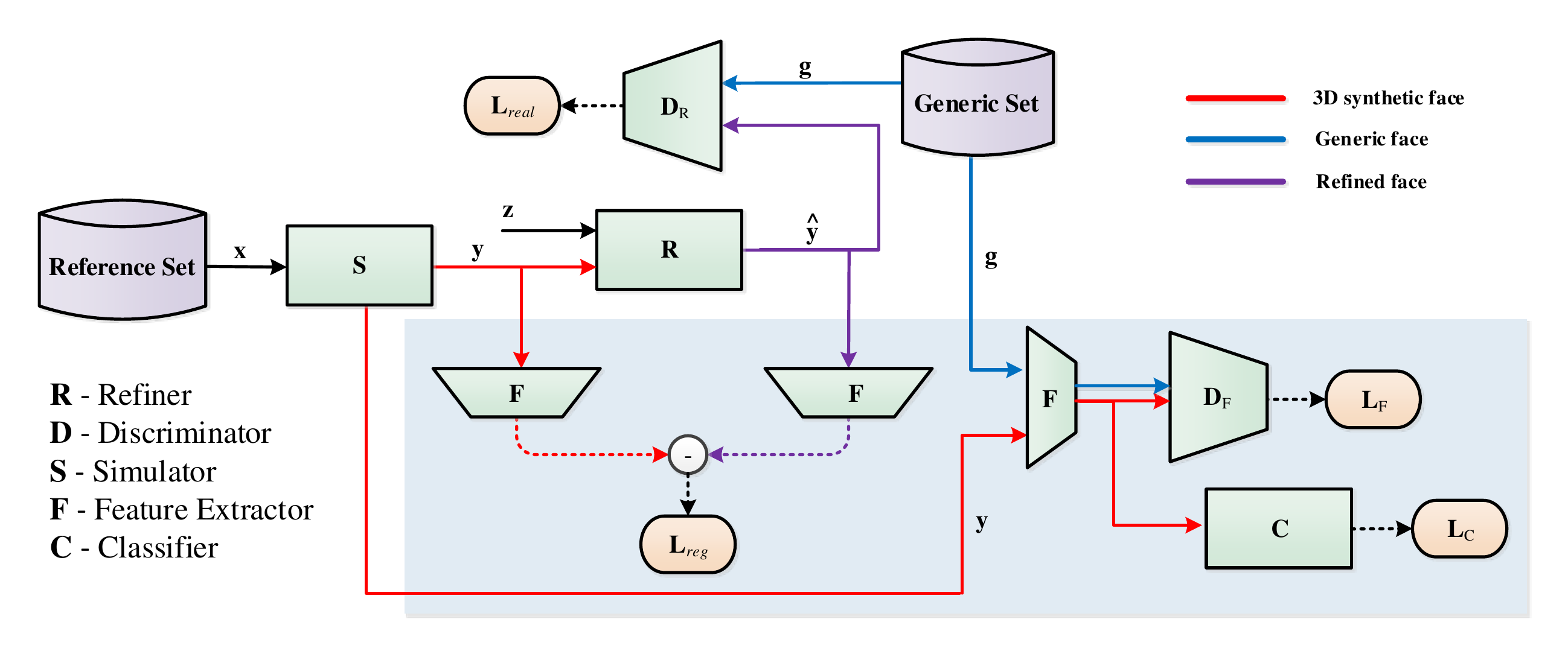}
\caption{\small Illustration of our proposed C-GAN architecture.
It incorporates a simulator $S$, a refiner $R$, two discriminators $D_{R}$ and $D_{F}$, and a classifier $C$ constrained by the simulated and noise vector. The blue box indicates the attributes preserving module.}
\label{fig:2}
\end{figure*}

Figure \ref{fig:2} depicts the overall C-GAN process within the approach for cross-domain face synthesis. Our approach is divided into three stages: (1) 3D simulation, (2) training the refiner, (3) refiner inference.  In the first stage, the 3D model of each reference still image is reconstructed via a 3D simulator and rendered under a specified pose. The rendered images, $\mathbf{Y}$, are then imported to the refiner (generator) to recover the information inherent in the target domain. In contrast to the vanilla GAN formulation \cite{goodfellow}, in which the generator is conditioned only on a noise vector, our model's generator is constrained on both a noise vector (z) and simulated image.

During the training stage, the refiner is trained to produce realistic images through an adversarial game with a discriminator network, $D_{R}$. The discriminator classifies a refined image as real/fake image. The refiner is further encouraged to generate realistic images while preserving the identity and capture conditions of $\mathbf{Y}$ by augmenting the refiner with a domain invariant feature extractor \cite{ganin}. The feature extracting is applied on both input and output of the refiner and the Euclidean distance of the two features is considered as an additional loss. The feature extractor $F$ must be invariant with respect to $\mathbf{Y}$ and $\mathbf{G}$ while including all identity and pose information. For this purpose, an additional adversarial game between another discriminator and the feature extractor is employed to train the feature extractor to be domain invariant. The second discriminator $D_{F}$ takes the output of the domain-invariant feature extractor and distinguish between the features extracted from the the refined images and real images. In order to guarantee that the extracted features include all the information of identity and pose, an identity-pose classifier predicts identity and pose of the refined images while being trained simultaneously on the labeled 3D simulated images, $\mathbf{Y}$. In this way, the target domain variations are effectively transferred onto the reference still images while specifying the pose shown in synthetic images, and without losing the consistency. 
The refiner in the proposed C-GAN shares ideas with methods for unsupervised domain adaptation \cite{ganin}, where labeled still images in the source domain and unlabeled video images from target domain are used to learn a domain-invariant embedding. We minimize the difference between the refined images and generic set while keeping the joint distribution information (on identity and pose). For stabilizing the training process of such dual-agent GAN model, we impose a boundary equilibrium regularization term. Once synthetic images are generated, any off-the-shelf classifier can be trained to perform the FR task.

\subsection{3D Simulator:}
\label{FS}
The simulated image set, $\mathbf{Y} \in \mathbb{R}^{d \times M }$, is formed by reconstructing the 3D face model of reference ROIs, $\mathbf{x}_{i}$, using a customized version of the 3DMM \cite{blanz1} in which the texture fitting of the original 3DMM is replaced with image mapping for simplicity \cite{fania}. The shape model is defined as a convex combination of shape vectors of a set of examples: 
\begin{equation}
\mathbf{s}= \mathbf{\bar{s}} + \sum_{k=1}^{m-1} \alpha_k . \mathbf{\hat{s}}_k~,
\label{eq11a}
\end{equation}
\noindent where $ \mathbf{ {\hat{s}}}_k$, $1 \le k \le m$ is the basis vector, $m$ is the number of the basis vectors, and $\alpha_k \in\mathbb [0,1]$ are the shape parameters.
The optimization algorithm presented in~\cite{blanz1} is employed to find optimal ${\alpha}_k^i$, for each $\mathbf{x}_i$. 
Then, the material layer of $\mathbf{x}_i$ is extracted and projected to the 3D geometry of $\mathbf{x}_i$. Given the 3D facial shape and texture, novel poses can be rendered under various pose by adjusting the parameters of a camera model. 
During the rendering procedure, the 3D face is projected onto the image plane with weak perspective projection:
%
\begin{equation}
\mathbf y^{j}= f*{\lambda}*\mathbf{R}^{j}*(\mathbf{\bar{s}} + \sum_{k=1}^{m-1} \alpha_k^i . \mathbf{\hat{s}}_k)+\mathbf{t}^{j}_{2d}~,
\label{eq0}
\end{equation}
\noindent where $\mathbf y^{j}$ is the $j^{th}$ reconstructed pose of $\mathbf x_i$, $f$ is the scale factor, $\lambda$ is the orthographic projection matrix
$\footnotesize\begin{pmatrix}
 1 & 0 & 0 \\ 
 0 & 1 & 0
\end{pmatrix}$,
$\mathbf{R}^{i}$ is the rotation matrix constructed from rotation angles and $\mathbf{t}_{2d}^j$ is the translation vector. 


\subsection{C-GAN Network Structure:}

The main part of C-GAN is the refiner ($R$) which improves the realism of a 3D simulator's output, $\mathbf{Y}$, using unlabeled images in the target domain, $\mathbf{G}$, while controlling their specific facial appearance (e.g. pose). The refiner, $R$, consists of an encoder $R_{enc}$ and a decoder $R_{dec}$. The encoder $R_{enc}$ aims to learn an identity and attribute representation from a face image $\mathbf{y}$: $\mathbf{f}(\mathbf{y})={R_{enc}}(\mathbf{y})$. The decoder ${R_{dec}}$ aims to synthesize a face image $\hat{\mathbf y}={R_{dec}}(\mathbf{f}(\mathbf{y}),\mathbf{z})$ where $\mathbf{z}\in \mathbb{R} ^{N^z}$ is the noise modeling other variance besides identity or attribute (e.g. pose). The goal of $R$ is to fool ${D_R}$ to classify $\hat{\mathbf{y}}$ as a generic image.

We adopt CASIA-Net \cite{casia} for $R_{enc}$ and $D_R$ where batch normalization and exponential linear unit are applied after each convolutional layer. A fully connected layer with logistic loss is added at the output of $D_R$. $R_{enc}$ and $R_{dec}$ are bridged by the to-be-learned identity representation $\mathbf{f}(\mathbf{y})\in \mathbb{R}^{320}$, which is the AvgPool output in $R_{enc}$ network. $\mathbf{f}(\mathbf{y})$ is concatenated with a random noise $\mathbf{z}$ and fed to $R_{dec}$. A series of fractionally-strided convolutions \cite{radford} transforms the ($320 + N_z$)-dim concatenated vector into a realistic image $\hat {\mathbf{y}}=R(\mathbf{y},\mathbf{z})$, which is the same size as $\mathbf{y}$.

An encoder, $F$, with the same structure as $R_{enc}$ is used at the input and output of $R$ to compare the domain invariant features of the simulated images to the refined images. As mentioned before, the discriminator, $D_F$, and classifier, $C$, are used to train $F$. Both $C$ and $D_F$ consist two fully-connected $1024$ unit layers. A fully-connected softmax loss for $k$ identity and $p$ pose classification is added to $C$ while a $1$ unit fully-connected logistic layer is added to $D_F$.
Table \ref{table0} shows the neural network structures of $R_{enc}$, $D_R$, $F$ and $R_{enc}$.


\begin{table}[t]
\renewcommand{\arraystretch}{1.3}
	\caption{The network structure of the proposed C-GAN architecture.}
	\vspace{.1cm}
	\label{table0}
	\centering
	\resizebox{\columnwidth}{!}{
     \begin{tabular}{ccccccc}
	\hline
	\multicolumn{3}{c}{ $R_{enc}$ and  $D_R$} & & \multicolumn{3}{c}{$R_{dec}$} \\ \cline{1-3} \cline{5-7} 
 Layer &   Filter/Stride  &  Output Size  & & Layer & Filter/Stride & Output Size \\ \hline
 &          &      &     &  FConv52 &  $3 \times 3/1$  & $6 \times 6 \times 320$ \\
Conv11 &     $3 \times 3/1$     &    $96 \times 96 \times 32$  &     &  FConv52 &  $3 \times 3/1$  & $6 \times 6 \times 160$ \\
Conv12 &     $3 \times 3/1$     &    $96 \times 96 \times 64$  &     &  FConv51 &  $3 \times 3/1$  & $6 \times 6 \times 256$ \\ \hline
Conv21 &     $3 \times 3/2$     &    $48 \times 48 \times 64$  &     &  FConv43 &  $3 \times 3/2$  & $12 \times 12 \times 256$ \\
Conv22 &     $3 \times 3/1$     &    $48 \times 48 \times 64$  &     &  FConv42 &  $3 \times 3/1$  & $12 \times 12 \times 128$ \\
Conv23 &     $3 \times 3/1$     &    $48 \times 48 \times 128$  &     &  FConv41 &  $3 \times 3/1$  & $12 \times 12 \times 192$ \\ \hline
Conv31 &     $3 \times 3/2$     &    $24 \times 24 \times 128$  &     &  FConv33 &  $3 \times 3/2$  & $24 \times 24 \times 192$ \\
Conv32 &     $3 \times 3/1$     &    $24 \times 24 \times 96$  &     &  FConv32 &  $3 \times 3/1$  & $24 \times 24 \times 96$ \\
Conv33 &     $3 \times 3/1$     &    $24 \times 24 \times 192$  &     &  FConv31 &  $3 \times 3/1$  & $24 \times 24 \times 128$ \\ \hline
Conv41 &     $3 \times 3/2$     &    $12 \times 12 \times 192$  &     &  FConv23 &  $3 \times 3/2$  & $48 \times 48 \times 128$ \\
Conv42 &     $3 \times 3/1$     &    $12 \times 12 \times 128$  &     &  FConv22 &  $3 \times 3/1$  & $48 \times 48 \times 64$ \\
Conv43 &     $3 \times 3/1$     &    $12 \times 12 \times 256$  &     &  FConv21 &  $3 \times 3/1$  & $48 \times 48 \times 64$ \\ \hline
Conv51 &     $3 \times 3/2$     &    $6 \times 6 \times 256$  &     &  FConv13 &  $3 \times 3/2$  & $96 \times 96 \times 64$ \\
Conv52 &     $3 \times 3/1$     &    $6 \times 6 \times 160$  &     &  FConv12 &  $3 \times 3/1$  & $96 \times 96 \times 32$ \\
Conv53 &     $3 \times 3/1$     &    $6 \times 6 \times 320$  &     &  FConv11 &  $3 \times 3/1$  & $96 \times 96 \times 1$ \\ \hline
AvgPool &     $6 \times 6/1$     &    $1 \times 1 \times 320$  &     &  &   & \\
\hline
FC ($D_R$ only) &          &    $1$  &     &  & \\ 
\hline
\end{tabular}
}
\end{table}

\subsection{Training the Refiner:}
Let a refined image be denoted by $\hat{\mathbf y}_i$, then $\mathbf{\hat{y}} = R({{\theta}_R};\mathbf{y})$  where ${\theta}_R$ is the function parameters. The key requirement is that the refined image, $\mathbf{\hat{y}}$, must look like a real generic image preserving the identity and pose information from the 3D simulator. To this end, $\theta_{R}$ is learned by minimizing a combination of two losses:
\begin{equation}
\mathcal{L}_{R}(\theta_R, \theta_F) = 
   \sum_{i} \mathcal{L}_{real} (\theta_R;{\mathbf{y}}_i, \mathbf{G}) + \lambda  \mathcal{L}_{reg} ({\theta}_F;{\mathbf{y}}_i)
\label{eq1}
\end{equation}
\noindent The first part of the cost, $\mathcal{L}_{real}$, adds realism to the simulated images, while the second part, $\mathcal{L}_{reg}$, preserves the identity and pose information.

The adversarial loss used for training the refiner network, $R$, is responsible for fooling $D_{R}$ into classifying the refined images as real. This problem is modeled a two-player minimax game, and update the refiner network, $R$, and the discriminator network.
$D_{R}$ updates its parameters by minimizing the following loss:
\begin{equation}
\mathcal{L}_D(\phi) = 
   - \sum_{i} \log(D_{R}(\phi;\hat{\mathbf{y}}_i)) - \sum_{j} \log(1 - D_{R}(\phi;\mathbf{g}_j))
\label{eq2}
\end{equation}
\noindent where $D_{R}(\cdot)$ is the probability of the input being a refined image, and $1-D_{R}(\cdot)$ that of a real one. 
For training this network, each mini-batch consists of randomly sampled $\mathbf{\hat{y_i}}$ and $\mathbf{g_i}$. 

The realism loss function employs the trained discriminator $D_{R}$ as follows:
\begin{equation}
\mathcal{L}_{real}({\theta}_R) =  \log(1 - D_{R}(R(\theta_{R};{\mathbf{y}}_i)))
\label{eq3}
\end{equation}
By minimizing this loss function, the refiner forces the discriminator to fail classifying the refined images as synthetic.

\noindent In order to preserve the annotation information of the 3D simulator, we use a self-regularization loss that minimizes difference between a feature transform of $\mathbf{Y}$ and $\mathbf{\hat{Y}}$, 
\begin{equation}
\mathcal{L}_{reg}({\theta}_F) =  \left\lVert F(\hat{\mathbf{y}}_i, \theta_F) - F({\mathbf{y}}_i, \theta_F)  \right\lVert
\label{eq1}
\end{equation}
\noindent where $F$ is the mapping from image space to a feature space, and $\| \cdot \|$ is the $\ell_2$ norm. 

Another adversarial game is employed to train the feature extractor network parameters (${\theta}_F$). For this purpose, the classifier, $C(.)$, assigns identity and pose information labels ($\mathbf{h}^i$) to a set of features extracted by $F$. In this way, $F$ learns to extract the features that are domain-invariant and consist information of identity and pose.
$F$ and $C$ are updated based on the identity and pose labels of $\mathbf{Y}$ in a traditional supervised manner. $F$ is also updated using the adversarial gradients from $D_F$ so that the feature learning and image generation processes co-occur smoothly.
\begin{equation}
\mathcal{L}_{C}(\theta_C)  = - \sum_i \sum_{j=1}^c \mathbf{h}_j^i ~ \text{log}(C(\theta_C;F(\hat{\mathbf{y}}_i)))
\end{equation}
\begin{equation}
\mathcal{L}_{D_{F}}(\gamma) = - \sum_{i} \log(D_{F}(\gamma;F(\hat{\mathbf{y}}_i))) - \sum_{i} \log(1 - D_{F}(\gamma;F(\mathbf{g}_j)))
\end{equation}
\noindent Given a realistic simulated images $\hat {\mathbf{y}}_i$ as input, $D_{F}$ outputs a binary distribution optimized by minimizing a binary cross entropy loss $\mathcal{L}_F$.The gradients are generated using the following loss functions:
\begin{equation}
 \mathcal{L}_{F}(\theta_F)  =  \sum_i \log(1 - D_{F}(F(\theta_F\hat{\mathbf {y}}_i)))
\end{equation}

\noindent where the $F$ and $D_F$ parameters are learned by minimizing $\mathcal{L}_F({\theta}_F)$ and $\mathcal{L}_{D_F}({\gamma})$ alternately. We leave  ${\gamma}$ fixed while updating the parameters of $F$, and we fix $\theta_F$ while updating $D_{F}$.

\section{Experimental Analysis} \label{EXP}


\subsection{Evaluation Methodology:}

The performance of the proposed and baseline methods was evaluated using two datasets for still-to-video FR. The \textbf{Chokepoint} \cite{Wong} consists of $25$ subjects walking through portal $1$ and $29$ subjects in portal $2$. Videos are recorded over $4$ sessions one month apart. An array of $3$ cameras are mounted above portal $1$ and portal $2$ that capture the entry of subjects during $4$ sessions. In total, the dataset consists of $54$ video sequences and $64,204$ face images. 
\textbf{COX-S2V} dataset \cite{Huang3} contains $1000$ individuals, with $1$ high-quality still image and $4$ low-resolution video sequences per individual simulating video surveillance scenario. The video frames are captured by $4$ cameras mounted at fixed locations. In each video, an individual walks through a designed S-shape route with changes in illumination, scale, and pose.

FR performance under SSPP scenario was assessed via the "recognition via generation" framework to validate our hypothesis that adding photo-realistic synthetic reference faces to the gallery set can address the visual domain shift, and accordingly improve the accuracy. Besides, since photographic results also indicate the performances qualitatively, the visual quality is also compared in our experiment.  We also compared our results with those obtained by flow-based Frontalization \cite{Hassner}.
During the enrollment of an individual to the system, $q$ simulated ROIs for each reference still ROI are generated under different poses using the conventional 3DMM \cite{blanz1}. The images are then refined using the controlled GAN that projects the capture conditions of the target domain on them while preserving their pose and identity. The gallery is formed using the original reference still ROIs along with the corresponding synthetic ROIs. During the operational phase, FR is performed using Siamese network model that is pre-trained using the VGG-Face2 dataset with Inception Resnet V1 architecture. The CNN feature extractors in this model is trained using stochastic gradient descent and AdaGrad with standard back-propagation \cite{facenet}. Finally, given the query (video) and reference (still) feature vectors, pair-wise matching is performed using the $k$-NN classifier based on Euclidean distance. 

In all experiments with Chokepoint and COX-S2V datasets, $5$ and $20$ target individuals are selected, respectively, to populate the watch-list, using one high-quality still image. Videos of $10$ individuals that are assumed to come from unlabeled persons are used as a generic set. The rest of videos including $10$ other unlabeled individuals and $5$ videos of the individuals who are already enrolled in the watch-list are used for testing. In order to obtain representative results, this process is repeated $5$ times with a different random selection of watch-lists and the average performance is reported with standard deviation over all the runs. The average performance of the proposed and baseline system for still-to-video FR is presented by measuring the partial area under ROC curve, pAUC(20\%) (using the AUC at $0 < FPR \leq 20$\%), and mean average precision, mAP. We further employed the Frechet Inception Distance (FID) \cite{heusel} to quantitatively verify the superiority of our synthetic faces.

\subsection{Results and Discussion:}

\begin{figure*}[ht]
\centering 
\advance\leftskip -0.2cm
\includegraphics[width=14cm]{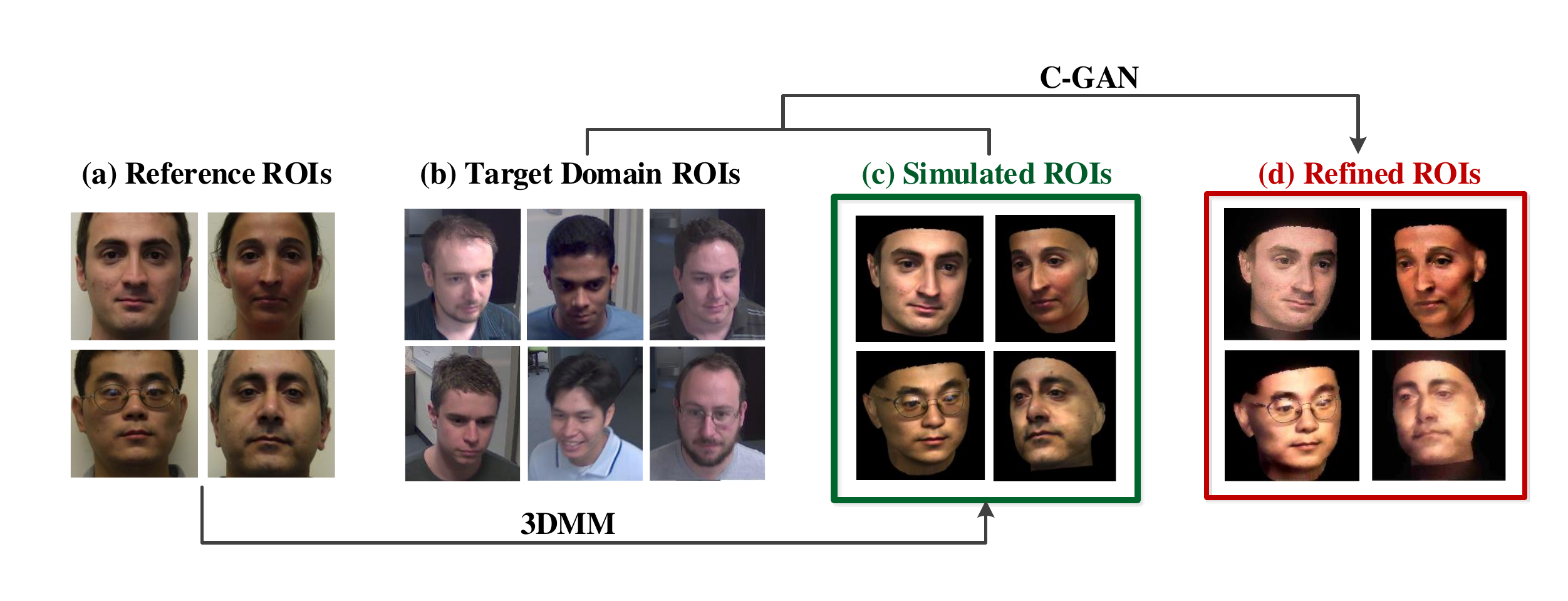}
\caption{\small Examples of the synthetic faces obtained with the proposed approach on Chokepoint database (ID\#1, ID\#5, ID\#6, ID\#16). The simulated images (c) are refined based on the target domain capture conditions (b) while preserving the identity of reference stills (a) under specific pose.}
\label{fig:3}
\end{figure*}

\begin{figure*}[b]
\centering 
\includegraphics[width=17cm]{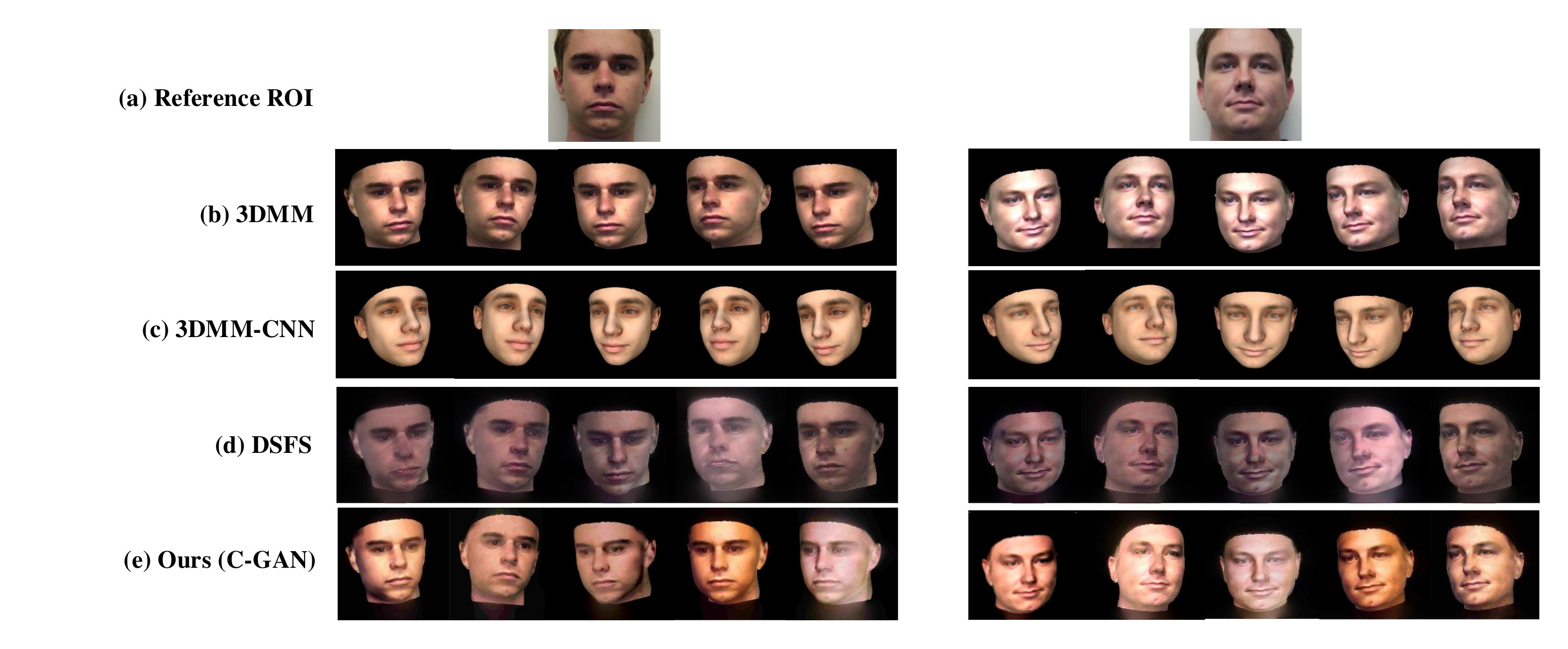}
\caption{\small Examples of facial images generated using state-of-the-art face synthesizing methods on Chokepoint dataset (ID\#23, ID\#25).}
\label{fig:31}
\end{figure*}

Figure \ref{fig:3} shows examples of the synthetic face images generated based on our proposed technique using the original reference still ROIs and generic set of the Chokepoint dataset. This show that the images refined using C-GAN can preserve their pose variations.
Figure \ref{fig:31} compares the qualitative results obtained with state-of-the-art techniques; (b) 3DMM \cite{blanz1}, (c) 3DMM-CNN \cite{Tran1}, (d) DSFS \cite{fania}, and (e) our proposed C-GAN.

Table \ref{table1} shows the average accuracy of a deep Siamese network for still-to-video FR that relies on the proposed and baseline methods for generating synthetic face images to augment the reference gallery. The baseline system is designed with an original reference still ROI alone. For our proposed C-GAN technique, the synthetic faces are generated with $5\degree$ step size within a range of $\pm5$ to $\pm60$ degrees in yaw, pitch, and roll. Consequently, we have $q=73$
synthetic face images in total for our experiments. For reference, the still-to-video FR system based on frontalization is also evaluated. Results indicate that by adding extra synthetic ROIs generated with C-GAN allows to outperform baseline systems. pAUC and mAP accuracy increases by about $3\%$, typically with $q=73$ synthetic pose ROIs for Chokepoint and COX-S2V datasets. Results suggest that that leveraging target domain information within the GAN framework while controlling its pose and identity can efficiently mitigate the impacts of the visual domain shift.

\begin{table*}
	\caption{\small Average pAUC and mAP accuracy of the Siamese network using the proposed and baseline methods for data augmentation. The '\# synth' columns show the minimum number of synthetic  samples needed to attain the highest level of accuracy.}
	\label{table1}
    \footnotesize
	\centering
	\begin{tabular}{l||ccc|ccc}
	\hline
    \multirow{2}{*}{\textbf{Techniques}} &  \multicolumn{3}{c}{\textbf{Chokepoint database}}   & \multicolumn{3}{c}{\textbf{COX-S2V database}} \\ \cline{2-7}  
	& \textbf{pAUC}(20\%) &  \textbf{mAP}  & \# \textbf{Synth} & \textbf{pAUC}(20\%) & \textbf{mAP} & \# \textbf{Synth}               \\ \hline \hline
Baseline                 & 0.908$\pm$0.018&0.861$\pm$0.020& N/A&   0.912$\pm$0.017&865$\pm$0.016& N/A\\
3DMM \cite{blanz1}       & 0.917$\pm$0.023&0.877$\pm$0.025&73&   0.928$\pm$0.026&872$\pm$0.027& 73\\
3DMM-CNN \cite{Tran1}    & 0.915$\pm$0.025&0.873$\pm$0.028&73&   0.922$\pm$0.024&871$\pm$0.028& 73\\
DSFS \footnote{This technique employs clustering to fond the optimal number of samples required for FR task.} \cite{fania}        & 0.923$\pm$0.018&0.880$\pm$0.019&17&   0.934$\pm$0.021&896$\pm$0.022& 14\\
SimGAN \cite{shrivastava}& 0.942$\pm$0.025&0.901$\pm$0.023&73&  0.948$\pm$0.023 &904$\pm$0.020& 73 \\
DR-GAN \cite{drgan}      & 0.931$\pm$0.019&0.893$\pm$0.017&73&  0.939$\pm$0.016 &903$\pm$0.017& 73\\
FaceID-GAN \cite{drgan}  & 0.936$\pm$0.023&0.905$\pm$0.019&73&  0.942$\pm$0.018 &911$\pm$0.022& 73\\
Frontalization \cite{Hassner}& 0.919$\pm$0.020 & 0.884$\pm$0.019&N/A&0.926$\pm$0.017 &892$\pm$0.020& N/A \\
C-GAN (Ours)             & 0.951$\pm$0.023&0.917$\pm$0.022&73&   0.957$\pm$0.019 &925$\pm$0.019& 73\\ \hline
\end{tabular}
\end{table*}

\begin{figure}[h!]
\vspace{.2cm}
        \centering       
        \subfigure[]{\includegraphics[width=40mm]{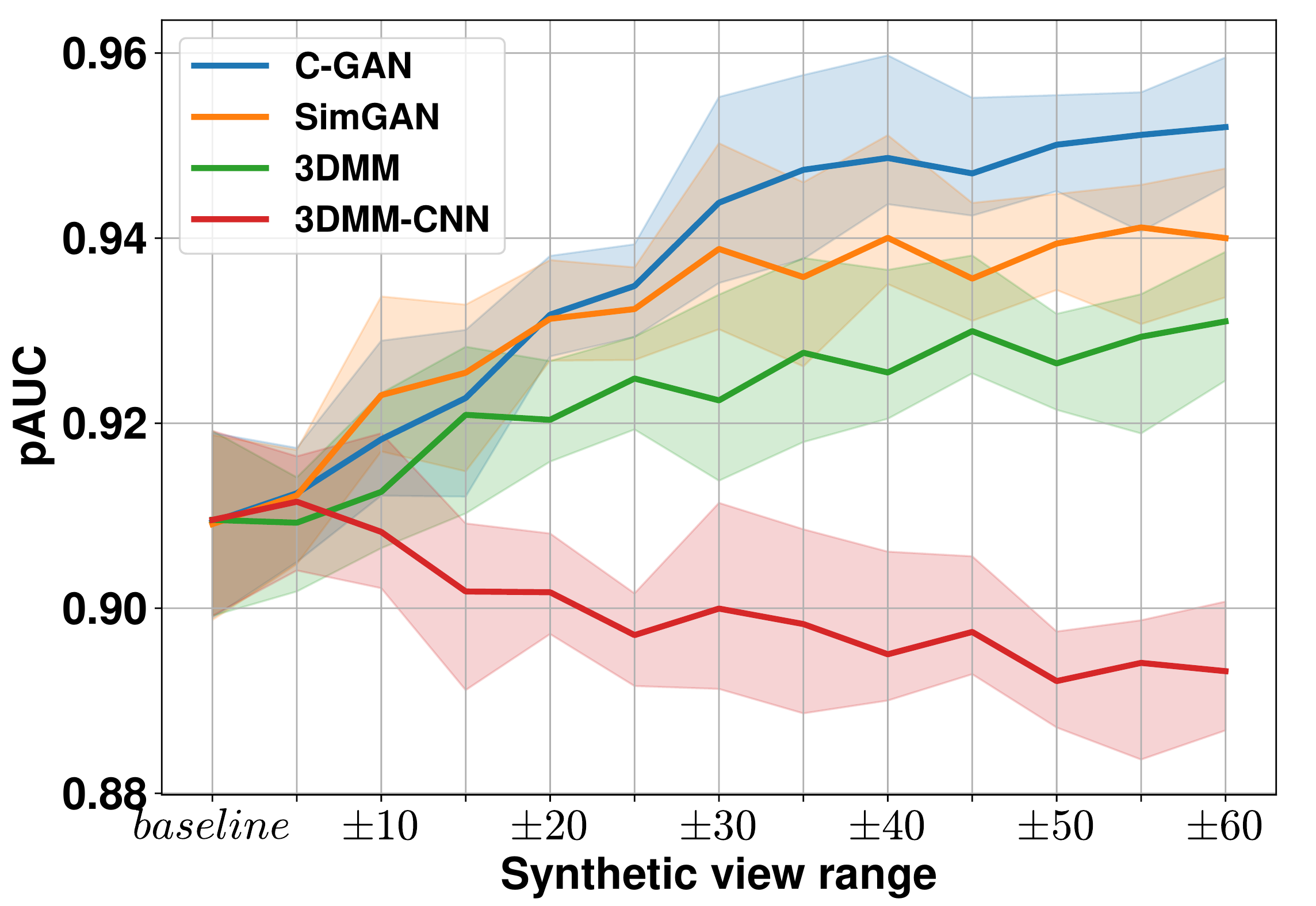}}
    ~ 
        \centering
        \subfigure[]{\includegraphics[width=40mm]{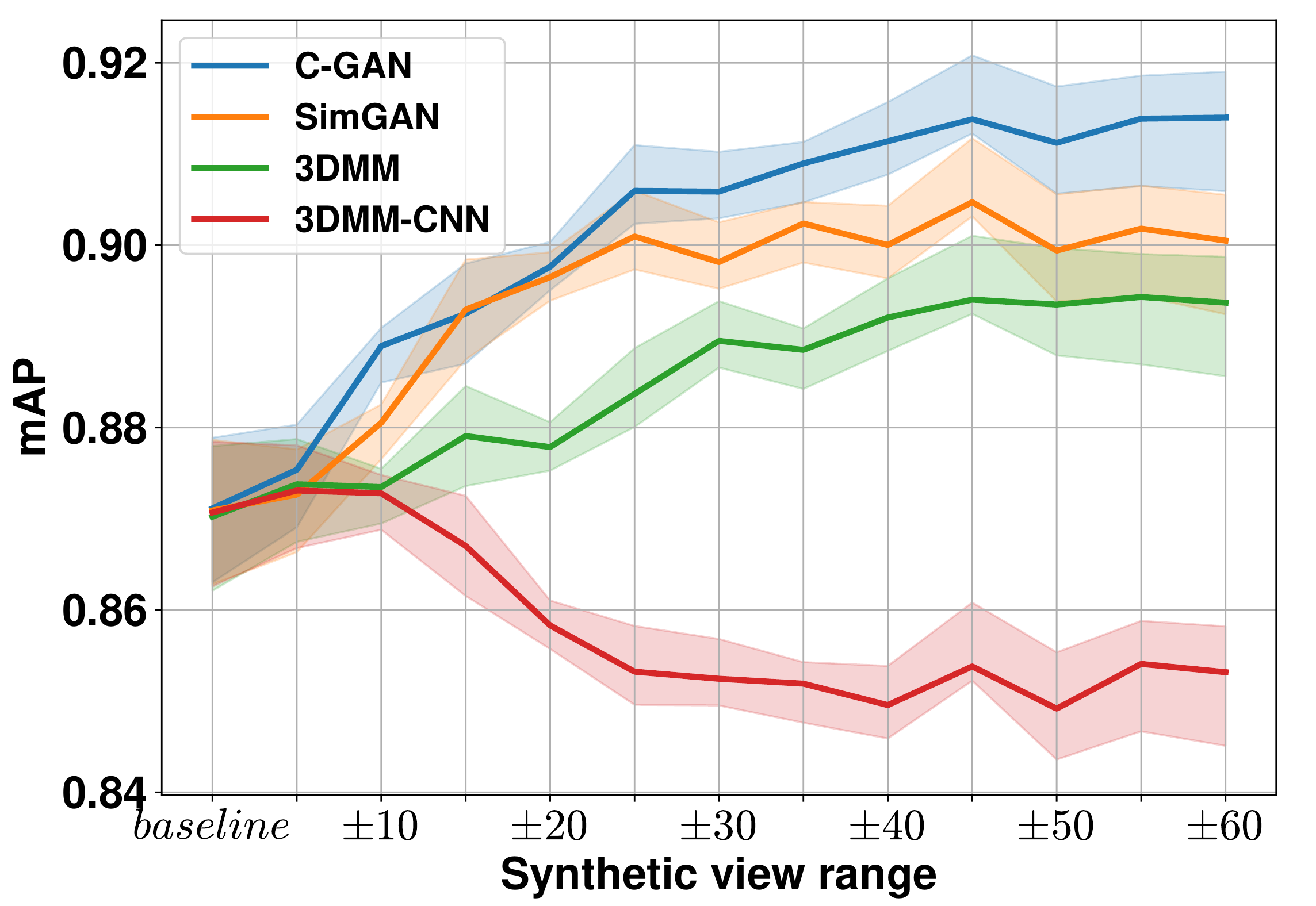}}
        \caption{\small Average pAUC(20\%) (a) and mAP (b) accuracy of the proposed and baseline techniques versus the size of the synthetic set on Chokepoint database. }
        \label{fig:5} 
\end{figure}

Figure~\ref{fig:5} shows the average pAUC(20\%) (a) and mAP (b) accuracy obtained for the implementation of still-to-video FR when increasing the number of synthetic ROIs per each individual. Adding synthetic ROIs generated under various capture conditions allows to significantly outperform the baseline system designed with the original reference still ROI alone. As shown in Figure \ref{fig:5}, accuracy trends to stabilize to its maximum value when the size of the synthetic faces is greater than  $q=73$ in C-GAN.  

Frechet Inception Distance (FID) \cite{heusel}\cite{cao} has been recently proposed to evaluate the performance of image synthesis tasks quantitatively where lower FID score indicates the smaller Wasserstein distance between two distributions.  Inception V3 model is employed to extract feature vectors from images.
Table \ref{table2} show the FID between the real and the synthesized faces across different yaw which demonstrates the effectiveness of our method.

\begin{table*}
\caption{\small FID across different views with Chokepoint and COX-S2V datasets.}
\label{table2}
\footnotesize
\centering
\begin{tabular}{l|ccl|cccc}
\hline
 \multirow{2}{*}{\textbf{Technique}}  & & \textbf{Chokepoint data}  & & & \textbf{COX-S2V data} \\ \cline{2-7}  
      & $\pm5^{\circ}$ & $\pm15^{\circ}$ & $\pm45^{\circ}$ & $\pm5^{\circ}$ & $\pm15^{\circ}$ & $\pm45^{\circ}$\\ \hline \hline  
3DMM \cite{blanz1}    & 22.3 & 23.4 & 25.7 & 20.5 & 21.4 & 21.7\\
3DMM-CNN \cite{Tran1} & 49.5 & 53.2 & 61.4 & 42.2 & 50.7 & 53.2\\
DSFS \cite{fania}     & 21.4 & 22.7 & 24.5 & 17.9 & 21.8 & 23.1\\
C-GAN (Ours)          & 20.9 & 22.1 & 23.8 & 17.3 & 20.9 & 21.5\\
\hline
\end{tabular}
\end{table*}

\begin{figure}[h!]
        \centering       
        \subfigure[]{\includegraphics[width=40mm]{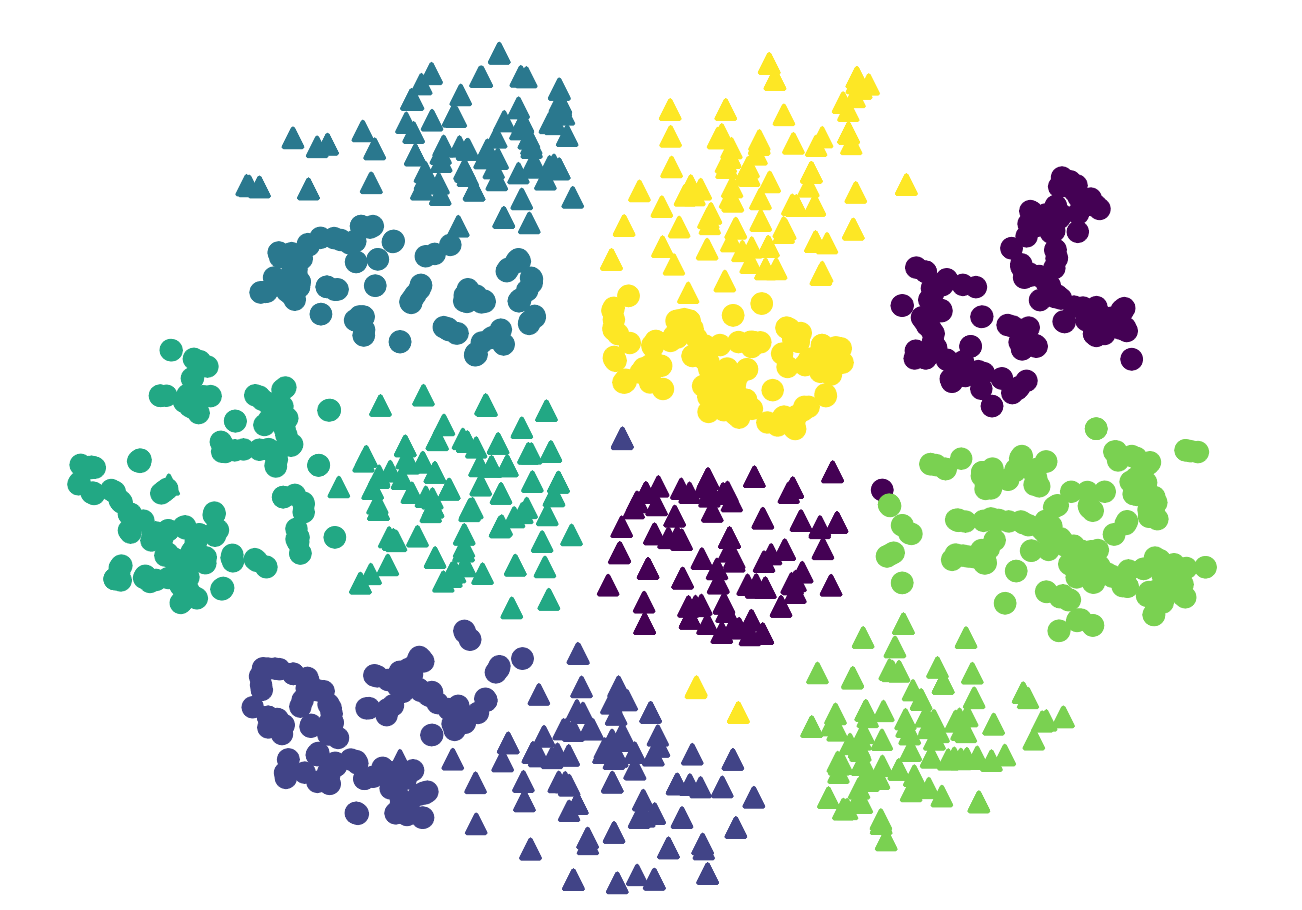}}
    ~ 
        \centering
        \subfigure[]{\includegraphics[width=40mm]{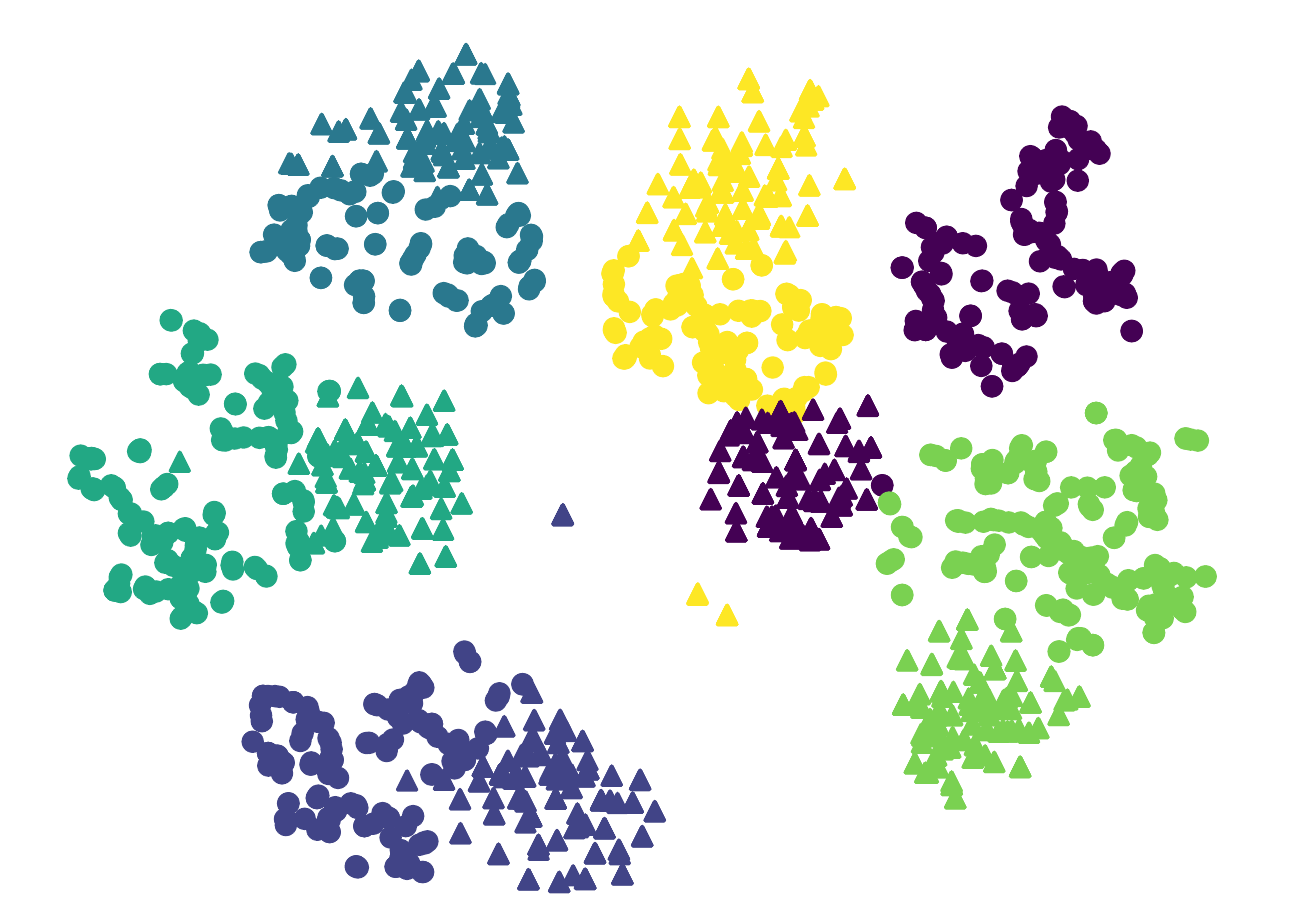}}
        \vspace{.1cm}
        \caption{\small t-SNE visualization.  Circles represent the generic set.  Triangles in (a) represent 3D simulated faces while triangles in (b) represent refined faces.}
        \label{fig:6} 
\end{figure}

To further evaluate the effectiveness of refiner in our C-GAN, we use t-SNE \cite{maaten} to visualize the deep features of simulated, refined and real faces in a 2D space. Figure \ref{fig:6} shows there is significant difference between the distribution of 3D simulated and real face. However, after refining the 3D simulated images, the distribution of the refined images become closer to the distribution of the real images.

\begin{figure}[ht]
        \centering       
        \subfigure{\includegraphics[width=40mm]{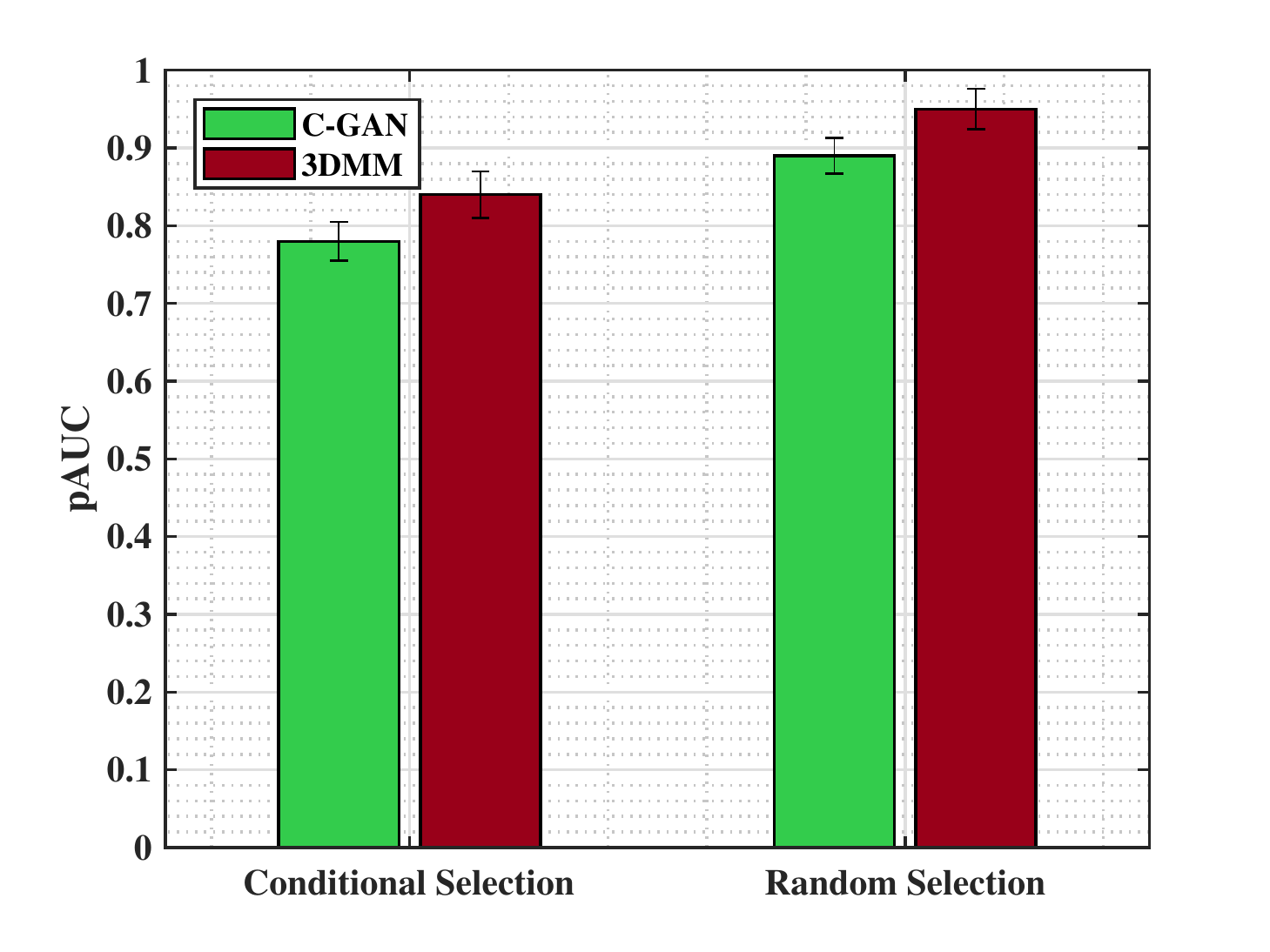}}
    ~ 
        \centering
        \subfigure{\includegraphics[width=40mm]{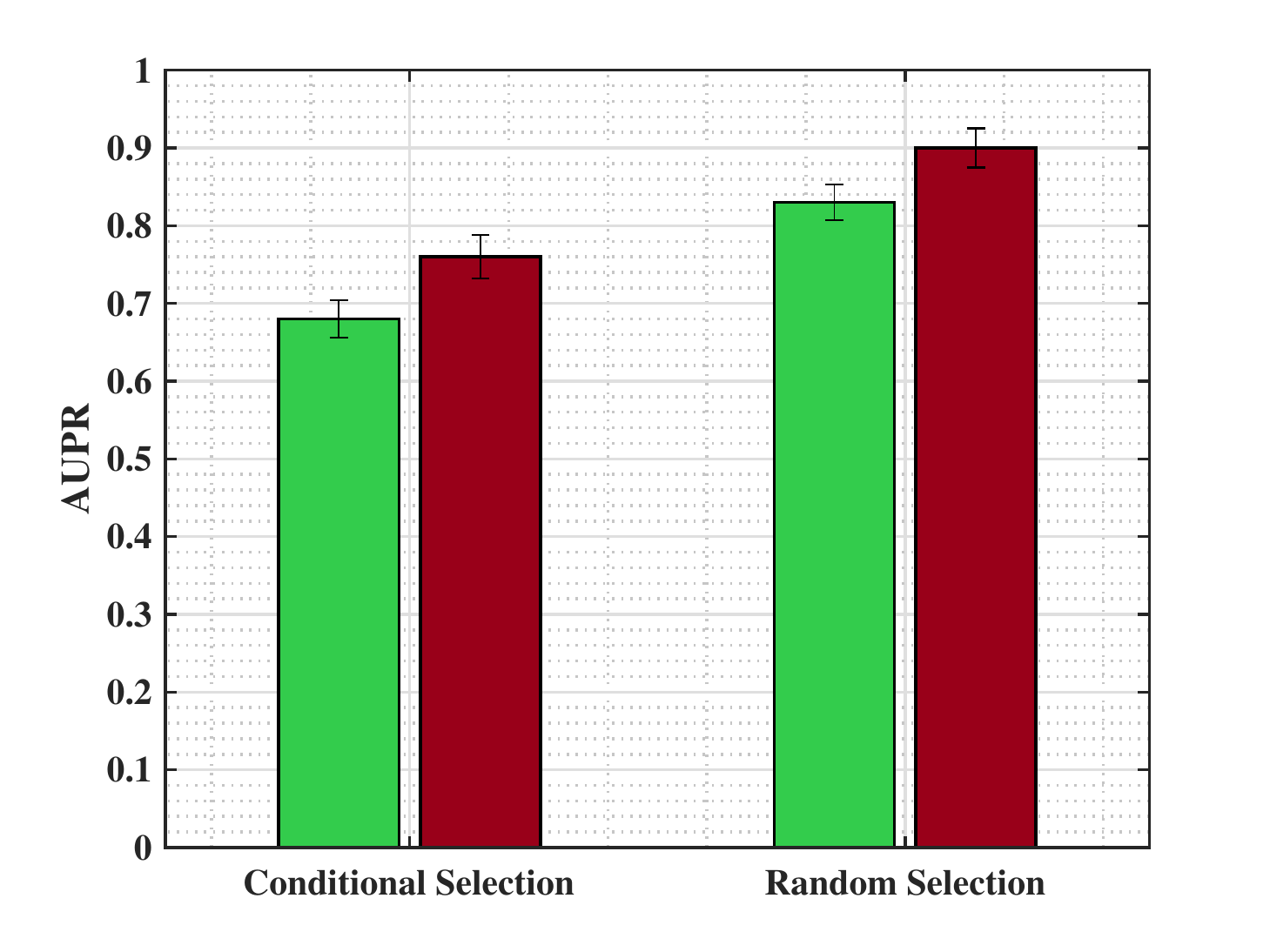}}
    ~ 
    
       \centering       
        \subfigure{\includegraphics[width=40mm]{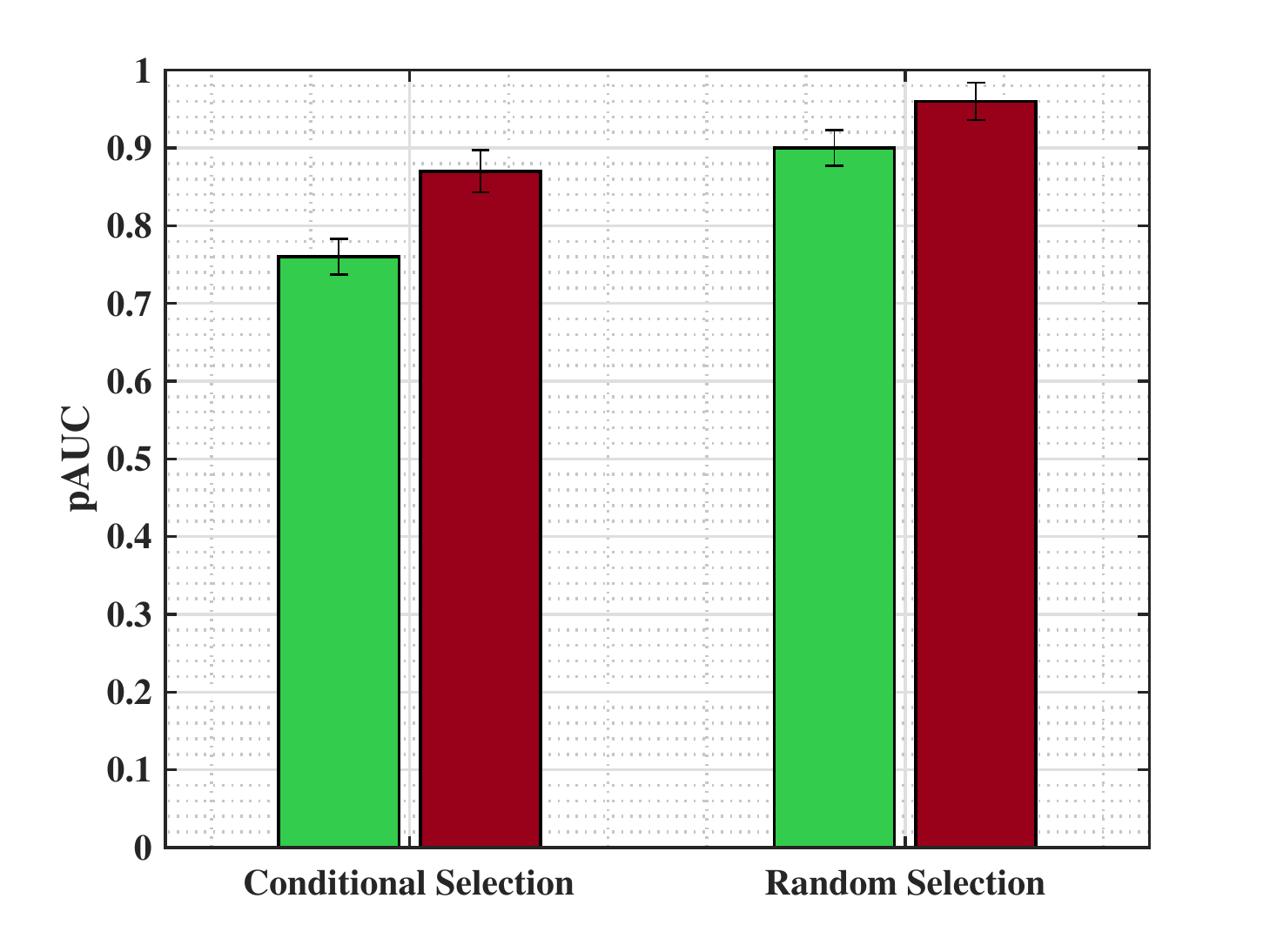}}
    ~ 
        \centering
        \subfigure{\includegraphics[width=40mm]{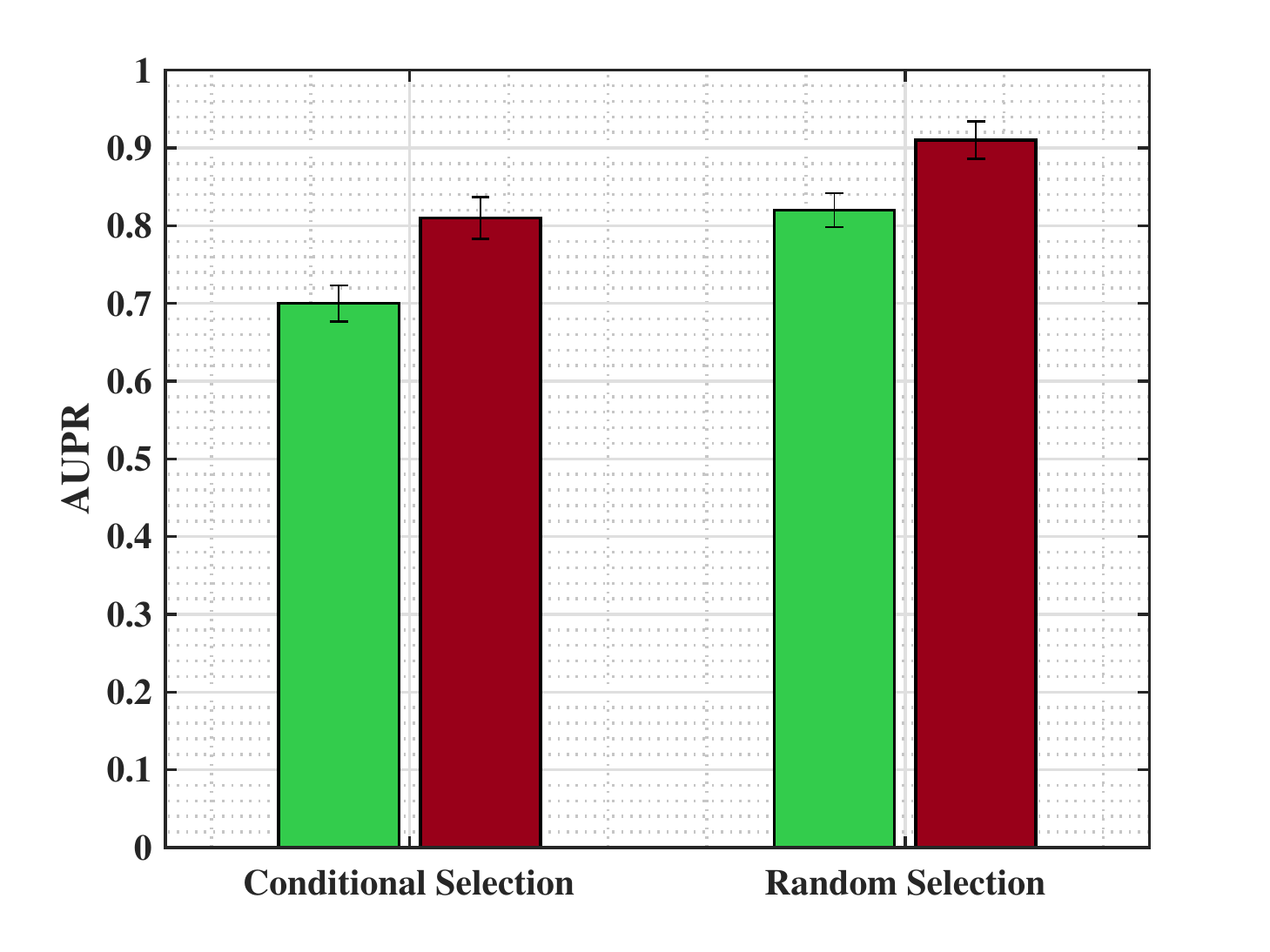}}
        \caption{\small Average pAUC(20\%) and AUPR accuracy for Siamese network with C-GAN and 3DMM face synthesis with $q$ specified and randomly selected faces on Chokepoint (a,b) and COX-S2V (c,d) databases. Error bars are standard deviation.\vspace{-.3cma}}
        \label{fig:7} 
\end{figure}

Figure \ref{fig:7} compares the performance of Siamese networks for FR when adding $73$  selected synthetic ROIs generated with the C-GAN versus $73$ randomly selected images (without condition). For reference, FR based on 3DMM face synthesizing is also evaluated.  Results in this figure show that the C-GAN with a specified range outperforms other models -- FR performance is higher when the gallery is designed using the representative views than based gallery comprised of randomly selected synthetic faces per person. The proposed C-GAN can therefore adequately generate representative facial ROIs for the reference gallery.

\subsubsection{Ablation Study:}
To evaluate the components of C-GAN ($D_F$ , $D_{R}$ , $C$), the model is trained by removing these modules while fixing the training process and all parameters. Recognition accuracy is evaluated on the synthetic images generated from each variant. We observe (Table \ref{table3}) that the accuracy will decrease by about $3\%$ if one module is not used. 

\begin{table*}
\caption{\small The results of ablation study with Chokepoint and COX-S2V datasets.}

\label{table3}
\footnotesize
\centering
\begin{tabular}{l|ccl|cccc}
\hline
\multirow{3}{*}{\textbf{Accuracy}} &  \multicolumn{6}{c}{\textbf{Removed Module}} \\ \cline{2-7} 
      & & \textbf{Chokepoint data}  & & & \textbf{COX-S2V data} \\ \cline{2-7}  
      & $D_{F}$ & $D_{R}$ & C & $D_{F}$ & $D_{R}$ & C\\ \hline \hline  
pAUC  & 0.905$\pm$0.022 & 0.901$\pm$0.023 & 0.882$\pm$0.020 & 0.908$\pm$0.021 & 0.902$\pm$0.025 & 0.891$\pm$0.027\\
mAP   & 0.873$\pm$0.024 & 0.868$\pm$0.021 & 0.854$\pm$0.019 & 0.885$\pm$0.018 & 0.875$\pm$0.020 & 0.859$\pm$0.024\\ \hline
\end{tabular}
\end{table*}

\vspace{-0.3cm}
\subsubsection{Time Complexity:}
Time complexity is estimated empirically, using the amount of time required to match $2$ facial ROIs with  given dataset. The average running time is measured with a randomly selected probe ROIs using a PC workstation with an Intel Core i7 CPU (3.41GHz) processor and 16GB RAM. 
Table \ref{table4} shows average matching time of deep Siamese networks over videos ROIs of the Chokepoint and COX-S2V datasets. The table shows time complexity grows the gallery size. The results suggest that the proposed approach represents an interesting trade-off between accuracy and complexity. 

\footnotetext{The DSFS technique employs a clustering on target capture conditions to find the optimal number of samples required for FR.}

\begin{table}
\caption{\small Average matching time over videos ROIs of the Chokepoint and COX-S2V datasets.}
\label{table4}
\footnotesize
\centering
\begin{tabular}{l||cc}
\hline
\multirow{2}{*}{\textbf{Techniques}} & \multicolumn{2}{c}{\textbf{Matching Time (sec)}}  \\ \cline{2-3} 
 		  & \textbf{Chokepoint}	 &  \textbf{COX-S2V} \\ \hline \hline  
Siamese Network {\cite{koch}}                 &         &  \\
\ $\cdot$ 1 frontal reference still / person  &  6.4    &  13.2  \\
\ $\cdot$ +73 uniform synthetic / person      &  129.7  &  186.1 \\
\ $\cdot$ +100 random synthetic / person      &  152.3  &  211.5 \\ 
\ Frontalization {\cite{Hassner}}             &  12.7    &  16.3 \\\hline 
\end{tabular}
\end{table}

\section{Conclusion}
\label{C}
In this paper, a cross-domain face synthesis approach with a new C-GAN model is proposed for data augmentation that generates highly consistent, realistic and identity preserving synthetic face images under specific pose conditions. The proposed model allows to mitigate the impact of some common issues with the original GAN model for data augmentation, such as lack of control and inconsistency. C-GAN leverages an additional adversarial game as third player to encourage the refiner during the inference to specify the capture conditions shown in synthetic images in a controllable manner. This allows augmenting to the gallery of a deep Siamese network with a diverse, yet compact set of synthetic views relevant to the target domain. Experimental results obtained using the Chokepoint and COX-S2V datasets suggest that the synthetic face images based on C-GAN allow us address visual domain shift, and thereby improve the accuracy of still-to-video FR system, with no need to generate a large number of synthetic face images. A future direction is to simulate and control other facial appearance (e.g. illumination and expression) during the face synthesis process. This can be further used to augment a dataset with representative images to train a deep neural network for still-to-video FR.

{\small
\bibliographystyle{ieee}
\bibliography{refs}
}

\end{document}